\RequirePackage{fix-cm}
\documentclass[journal]{IEEEtran}
\IEEEoverridecommandlockouts                              % This command is only needed if 

\usepackage{times} % assumes new font selection scheme installed
\usepackage{cite}
\usepackage{url}
\usepackage{amsmath}
\usepackage{amssymb}
\usepackage[table]{xcolor}

\newcommand{\matr}[1]{#1}

\usepackage[pdftex]{graphicx}
\graphicspath{{"./Figures/"}}
\usepackage{epstopdf}
\DeclareGraphicsExtensions{.pdf,.jpg,.png,.tif}
\usepackage{subcaption}
\title{The Use of Gaze-Derived Confidence of Inferred Operator Intent in Adjusting Safety-Conscious Haptic Assistance}

\begin{document}

\author{Jeremy D. Webb, Michael Bowman, Songpo Li, and Xiaoli Zhang$^{*}$,~\IEEEmembership{Member,~IEEE}% <-this % stops a space
\thanks{Jeremy D. Webb is an engineer with Apple Inc.(e-mail: jewebb@mines.edu.)}
\thanks{Michael Bowman is a Postdoctoral Fellow in Perelman School of Medicine at University of Pennsylvania, Philadelphia, PA 19096 USA (e-mail: michael.bowman@pennmedicine.upenn.edu).}
\thanks{Songpo Li is an Associate Scientist with Honda Research Institute US (e-mail: songpo\textunderscore li@honda-ri.com).}% <-this % stops a space
\thanks{Xiaoli Zhang is an Associate Professor in the Department of Mechanical Engineering at Colorado School of Mines, Golden, CO 80401 USA ($^{*}$corresponding author, phone: 303-384-2343; fax: 303-273-3602; email: xlzhang@mines.edu).}
}
\maketitle
\thispagestyle{empty}
\pagestyle{empty}
%%%%%%%%%%%%%%%%%%%%%%%%%%%%%%%%%%%%%%%%%%%%%%%%%%%%%%%%%%%%%%%%%%%%%%%%%%%%%%%%
\begin{abstract}
Humans directly completing tasks in dangerous or hazardous conditions is not always possible where these tasks are increasingly be performed remotely by teleoperated robots. However, teleoperation is difficult since the operator feels a disconnect with the robot caused by missing feedback from several senses, including touch, and the lack of depth in the video feedback presented to the operator. To overcome this problem, the proposed system actively infers the operator's intent and provides assistance based on the predicted intent. Furthermore, a novel method of calculating confidence in the inferred intent modifies the human-in-the-loop control. The operator's gaze is employed to intuitively indicate the target before the manipulation with the robot begins. A potential field method is used to provide a guiding force towards the intended target, and a safety boundary reduces risk of damage. Modifying these assistances based on the confidence level in the operator's intent makes the control more natural, and gives the robot an intuitive understanding of its human master. Initial validation results show the ability of the system to improve accuracy, execution time, and reduce operator error.
\end{abstract}

%%%%%%%%%%%%%%%%%%%%%%%%%%%%%%%%%%%%%%%%%%%%%%%%%%%%%%%%%%%%%%%%%%%%%%%%%%%%%%%%
\section{Introduction}
\label{sec:intro}

\subsection{Context and Motivation}

Remotely operated robotic procedures performed has continued to increase each year. These procedures include scenarios where a human may find it difficult to achieve a task, such as telesurgery, and environments where it is dangerous for a human to be present such as in environments contaminated by chemical, radioactive, or explosive hazards. The environmental challenges where these robotic systems are used include bomb disposal/mine clearing robots ~\cite{usingrobotsinhazardous}, robots for making repairs in space, robots for handling nuclear material~\cite{chemicalandhazardous}, hazardous waste handling robots~\cite{internationalelectrotechnical}.

Use of these robots greatly improves the safety and comfort of the humans performing the tasks, yet also adds complexity and difficulty in achieving the goal. The reason for the difficulty is mainly due to the “disembodiment” problem ~\cite{contributionofnueroscience}. Where this problem describes the fact that the operator is not physically performing the tasks in the environment yet must mentally accomplish the task. The correspondence issue is inherent in the system as the operator lacks sufficient sensory feedback. Specifically, in a typical setup, an operator views a screen and controls a robot with a joystick; however, feedback from touch and sound as well as depth into the screen are all missing. Not being able to distinguish the Z-order of objects can create erroneous complications. For example, in surgery, unexpected tissue damage, longer operating times, and increased stress for the surgeon can all result from a lack of depth information. Likewise, for a bomb disposal robot, errors caused by the operator’s unclear understanding in depth could result in a bomb denotating prematurely. It also has been shown that depth perception is paramount in successfully performing grasping tasks for human’s using their own hands, where performance degrades as the depth perception becomes inaccurate ~\cite{geometriccomputations}. Furthermore, trying to determine the depth of an object can distract a teleoperator. Since the operator is not using their own arm to complete the task, they will not have a good intuitive understanding of the dynamic behavior of the robot being controlled. These issues can cause mistakes that lead to unintentionally harming the surrounding environment which has the potential to be far more costly than failing to complete the teleoperation task.

Existing interfaces for teleoperated robots attempt to solve these issues in various ways, but many are difficult to operate. One interface uses fixed targets as reference points and an oscillating camera towards and away from these points to give an operator a better sense of the environment~\cite{usingradialoutflow}.The largest problem with this approach is the difficulty for the operator to accomplish their task with a constantly moving camera. Another suggested method to help the operator understand depth in a teleoperation scenario is to reconstruct a virtual 3D environment by using stereoscopic video~\cite{skillacquisition,medicalimaging}. Other solutions to aid the operator determine depth is to change the lighting conditions of the environment and provide visual cues~\cite{operatorvisionaids},] and provide a target object’s pose using machine vision~\cite{machinevison}. It should be noted this last method will fail whenever new objects are encountered.  Another approach is to immerse the operator in the environment by using a head-mounted display, which allows the user to look around naturally~\cite{immersive3Dteleoperation}, but this requires more complex equipment.

There is therefore a need for an intuitive control interface that can restore some of the sensory feedback lost during teleoperation and increase the accuracy of task completion. Such a system should take into account the operator's intent to provide accurate assistance for real-time control applications, and cooperate in a way that is comfortable for the operator~\cite{attentionaware}. In this case, the operator's gaze can serve to indicate their intent, or the final goal of a manipulation, and this information can be used to guide the operator's hand to the target. This can be accomplished through the use of haptic forces, which function as a partial restoration of sensory feedback. Furthermore, the provided assistance should be adjusted based on the system's confidence in the operator's intent~\cite{teleoperationwithintelligent}. Since the system cannot be 100\% sure of the predicted operator's intent, the confidence level in the intent should be used to moderate the strength of the provided haptic forces. This will ensure the system is robust and provides accurate assistance.

The proposed system takes advantage of the natural visuomotor behavior of human beings. Several human visuomotor and cognitive behavior studies indicate that one's gaze leads their hands during execution of a grasping or reaching task. Specifically, when a human decides to pick up an object, he first looks at the object, then focuses on the part of the object to be grasped, and finally executes the reaching movement. Typically, the gaze fixates on an object before interacting with it and stays fixed on the object until the task is completed~\cite{oculomotorbehavior, inwhatwaysdoeye}. The average lead time for a grasping task has been found to be 3 seconds~\cite{lookaheadfixations}. A human’s eye gaze has been shown to focus on certain parts of an object depending on the current task~\cite{goaloriented}, however, initially the gaze is focused on the object’s center~\cite{theroleofobject}. Furthermore, in a comprehensive review~\cite{spatialtransformations} describes how the human brain maintains a model of the “eye-head-shoulder system” and treats gaze as a feedforward mechanism when reaching for an object.  Even on 2D displays, eye-movements indicate the user’s intention and thus, can be used for “highly-intuitive” computers~\cite{gazestrategies}. ]. These anticipatory fixations also happen in teleoperation~\cite{remotecontrol}. Therefore, incorporating the operator's gaze into the teleoperation control interface to specify the center of the haptic assistance is a natural extension of normal human behavior. There is a need for such a system, as described by \cite{humanorientedcontrol} which speaks of the need for human perception models in haptic teleoperation to improve human-in-the-loop control.

\subsection{Previous Work}

Incorporating haptic feedback into a teleoperation system can restore some of the sensory feedback that is lacking. Haptic feedback, in this case, refers to applying forces to the operator that are dependent on the system's state. Using haptic feedback in teleoperation has been explored in a variety of studies, especially in remote surgery applications. Researchers have shown that using a ``computerized force feedback endoscopic surgical grasper'' in minimally invasive surgery leads to significant performance gains over using a regular endoscopic grasper \cite{forcecontrolledendoscopic}. Similarly, using force feedback in blunt dissection reduces tissue damage and the force used in robotic surgery \cite{theroleofforcefeedbackinsurgery}. Haptic feedback has been shown to improve teleoperation control in general, as well. One study investigated using haptic feedback for training one's hands to follow a certain trajectory, demonstrating that haptic feedback does improve the training \cite{hapticguidance}. This indicates that haptics can be used to teach a more straightforward path to the goal. Other studies have shown the use of potential fields in haptics to guide the operator by pushing their hand away from objects and/or towards the goal (termed guidance virtual fixtures) \cite{hapticvirtualfixtures, hapticallyaugmented, gazecontingentmotor}. One such project developed a potential field to control unmanned aerial vehicles \cite{artificialforcefield}.

Often, providing force feedback in terms of a virtual fixture only solves one issue with teleoperation, accurately reaching the target. For many teleoperation tasks, such as minimally invasive surgery (MIS), some damage to the surrounding environment is unavoidable~\cite{invivosofttissue}. However, this damage can be minimized by ensuring the operator only moves the robot in allowable regions. This is enforceable using haptic forbidden-region virtual fixtures. A number of forbidden-region virtual fixtures have been demonstrated in various research, especially those concerned with MIS. One shows that forbidden-region virtual fixtures that move with a portion of the environment, such as a beating heart, can increase user precision~\cite{designconsiderations}. A method which assists the user performing MIS by placing conical forbidden-region virtual fixtures at a set of predetermined locations has also been developed~\cite{gazecontingentmotor}. Others have built methods to automatically generate forbidden-region virtual fixtures based on the output of RGB-D cameras to protect sensitive areas~\cite{acomputervisionapproach,usingkinectandahaptic}.

The drawback to the approaches described above is that they rely mainly on situational context alone to determine how to implement the virtual fixtures. Instead, as noted earlier, the system should incorporate the operator's intent and confidence in the prediction of that intent into the control loop to provide intuitive and accurate assistance. One way to do this is to use the operator's gaze.

Gaze as a control input has been used in a variety of assistive mechanisms to help the teleoperator visualize the robot workspace~\cite{towardremote,gazecontingent,areviewof,attentionaware}, and to direct robot navigation. In particular, researchers have demonstrated successful use of gaze gestures to control teleoperated drones~\cite{humanrobotinteraction}, and gaze contingent regions, or ``hot-spots'', have been used to specify a robot's direction of movement~\cite{telegaze,mobilerobotnavigation}. These approaches do not use the operator's gaze as an indication of intent, instead the user must consciously focus on a particular area to provide input for the signal. This can cause fatigue for the operator and distract them from completing the goal.

Using the operator's gaze to infer their intent can provide a more natural control scheme.  This is demonstrated in ``predicting a driver's intent to change lanes'', which used head motion and eye data to train a discriminative classifier to perform the prediction~\cite{ontherolesofeyegazeandhead}. As explained, use of the user's intent has been shown in other applications, but using intent prediction in haptics is a new area that has not been explored. Additionally, the shortcoming in the previous approaches are that they provide only a binary output. For example, the user intends to change lanes, or does not intend to. In this case, an important aspect of the system's ability to make decisions has been left out: the probability that the predicted intent is correct. This component is necessary to ensure robust control and decision making. For example, undesired behavior could be encountered if the system attempts to assist with lane changing when the intent prediction is hovering between intent and no intent. Instead the system should provide assistance based on its confidence in the predicted intent. Another reason for this is because the user's gaze is really an observation mechanism, not a control input (known as the Midas Touch problem \cite{midastouch}). Therefore, to reduce inference error the system should take into account the likelihood that the user actually has an intent.

\subsection{Research Contributions}

The proposed system aims to reduce risk and enhance performance in realtime teleoperation through three approaches: gently guide the operator's hand toward the goal point; prevent unwanted destruction of the surrounding teleoperation environment; and ensure control is natural and intuitive by modifying the previous two approaches based on the system's confidence in the operator's inferred intent. The intent inference restores a teleoperator's eye-hand coordination through incorporation of the operator's natural visuomotor behavior by using their gaze to determine the reaching target before the process begins. A force then gently pushes the operator's hand towards the target. Simultaneously, a safety boundary prevents harm to the environment by restricting joystick movement to a small area around the target point. The size and strength of both of these virtual fixtures is adjusted based on the system's confidence in the inferred intent and the specific task. As discussed in~\cite{asurveyofenvironment}, including awareness of the environment and task in a teleoperation control scheme can give great improvements in performance. Additionally, the biggest challenge when using virtual fixtures is determining the appropriate strength of the fixture~\cite{asurveyofenvironment}. As noted, this system deals with this issue by dynamically assigning the strength based on the current situation and the probability that the predicted operator's intent is correct.

The contributions of this system include:
\begin{enumerate}
	\item development of haptic virtual fixtures which are based on the operator's inferred intent to ensure control is instinctive and improves performance
	\item a novel gaze-driven method for determining the level of confidence in the predicted human intent
	\item real time adjustment of the haptic virtual fixtures based on the operator's predicted intent and confidence in that prediction, which reduces risk and increases success rate in teleoperated tasks
	\item evaluation of the effectiveness of the intent-driven haptic assistance with confidence adjustment
\end{enumerate}

The operator’s gaze is used to indicate the final goal of the joystick motion because, for executing grasping tasks, the hand of the operator follows their gaze. Additionally, the confidence in the intent is computed using features inherent in the operator's gaze. Using this approach, the system can predict the operator’s intention and determine its own confidence in the prediction without the need for extra effort by the operator. Adjusting the strength of guidance and safety boundary based on the system's confidence in the inferred intent allows the user to teleoperate the robot as normal. The shape of the safety boundary is chosen to minimize the risk associated with completing a given task. The guidance force is computed using a potential field method. Using a potential field allows the spatial uncertainty in the predicted target location to be taken into account. Errors in gaze tracking, target location determination, and robot end-effector location all contribute to the spatial uncertainty.

The proposed system will increase precision, safety, and ease the use of teleoperation, thus improve task performance by reducing the time it takes to complete a task and increase the comfort of remote operators.

\section{System Overview}
\label{sec:system_architecture}
\begin{figure}[!t]
	\centering
	\includegraphics[width=\linewidth]{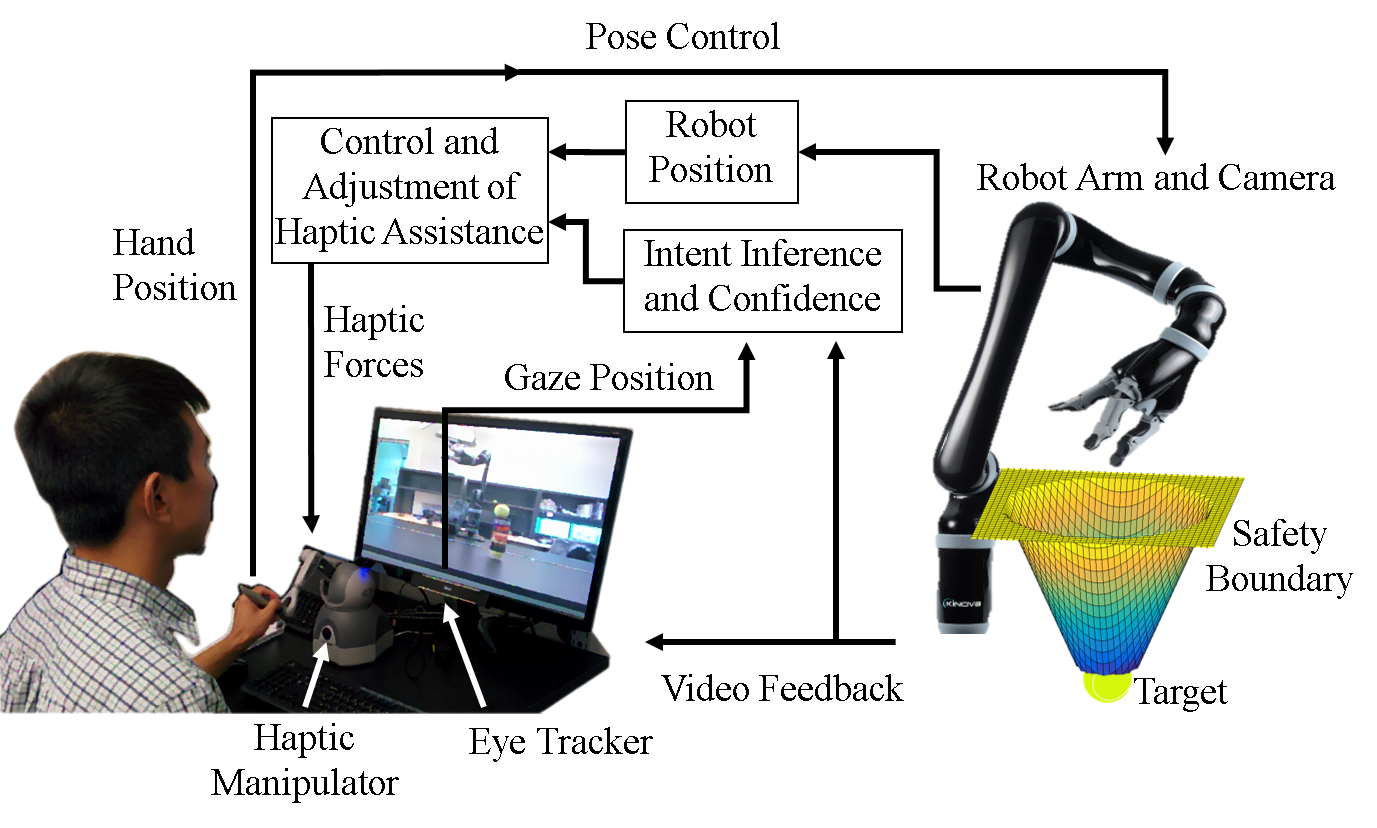}
	\caption{System overview}
	\label{fig:fig_overall}
\end{figure}

The overall system predicts the goal position from the operator's gaze, determines the confidence in the prediction, guides the operator's hand to the target using a force based on the output of the potential hybrid control, and places a safety boundary (forbidden-region virtual fixture) around the goal point. In this system, shown in  in Fig. \ref{fig:fig_overall}, the video feedback shows the operator’s gaze location, which is the goal position, while a 6 DOF joystick gives the operator the manual control input for orientation and position. The guidance force and the safety boundary is provided by the joystick, which is also a haptic device. Most of the time, the operator should not feel the boundary at all. Their hand will only come into contact with it if they attempt to command the robot to a position too far from the target.

\begin{figure}
	\centering
	\includegraphics[width=\linewidth]{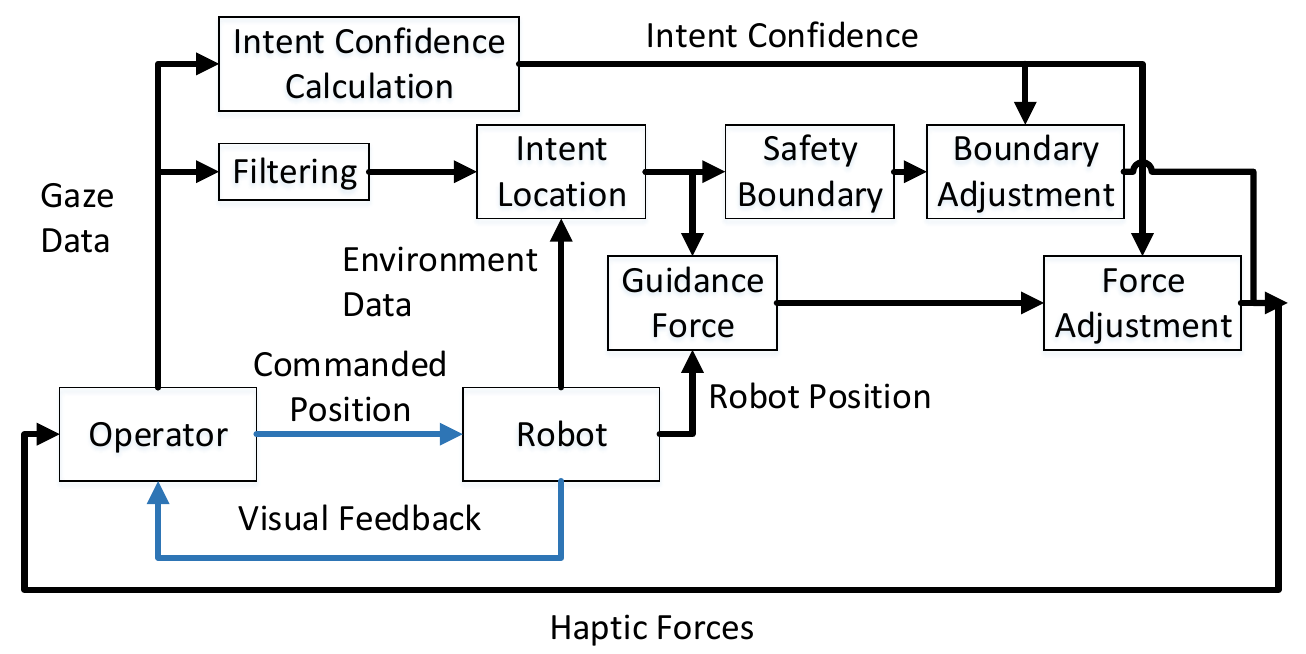}
	\caption{Overall control flow which illustrates the role of the operator's gaze in the hybrid joystick-gaze control method. The blue lines represent the control flow for traditional teleoperation.}
	\label{fig:control_flow}
%	\vspace{-0.5cm}
\end{figure}

The overall control flow with a more detailed view is shown in Fig. \ref{fig:control_flow}. The operator’s fixation location is determined by acquiring data from the eye tracker and filtering it as further discussed in section \ref{sec:gaze_details}. The confidence in the operator's intent is also calculated using the gaze data. By combining the gaze information from the robot environment, a fully specified spatial position of the target is determined. The safety boundary is placed with its center at the target position and its parameters are adjusted based on the confidence level of the intent. Simultaneously, the target position is blended with the position of the joystick using the novel potential field method to determine the force to apply to the operator's hand. The pose of the joystick is then fed into the controller for the robot.

As shown in Fig. \ref{fig:control_flow}, this system is considered a closed loop through the user viewing video feedback from the robot and adjusting the joystick position or orientation accordingly. Additionally, the guidance force pushes the operator's hand towards the target. The operator's gaze assists with this because it incorporates the operator's intention into the control by indicating the target position. In summary, the operator’s gaze location indicates their intended target and the system assists them in reaching this target by actively guiding their hand towards this position.

\section{Motion Intent Extraction from Eye Movement}
\label{sec:gaze_details}

Humans eyes naturally make involuntary movements and motions such as blinking, rolling, and microsaccades. Therefore, it is necessary for a method to filter the raw gaze data to determine the operator’s fixation location. The filter which was chosen to combat this was an adaptive-length sliding window~\cite{attentionaware}.

When gaze is used to control a system, it becomes necessary to determine a way to distinguish an intentional command from an unintentional one. Because the gaze is always “active”, this complicates the problem which is referred to as the Midas touch problem~\cite{midastouch}. One approach to overcoming this distinction is to use the dwell time method. This method considers a command to be confirmed when the gaze stays on a location for a set amount of time. Alternatively, an option could be to require a certain number of blinks to confirm a command. In this system, a method of determining the likelihood that a predicted intention is correct has been developed. The following section describes the approach.

\begin{figure}
	\centering
	\includegraphics[width=0.7\linewidth]{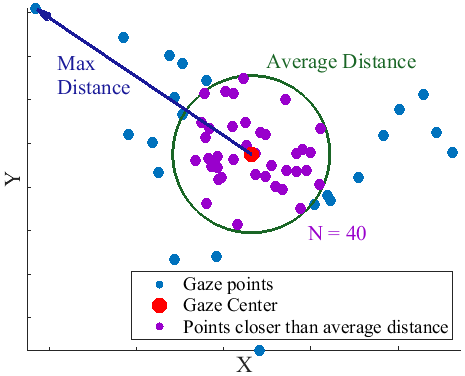}
	\caption{Example of the gaze features used to determine the intent confidence. X and Y are the screen coordinates in pixels.}
	\label{fig:gaze_features}
	%	\vspace{-0.5cm}
\end{figure}

The confidence in the predicted intent derived from the operator's gaze is determined by using a naive Bayes classifier fitted to three processed gaze features. These features are: the maximum euclidean distance of the gaze points to the gaze center, the average distance to the gaze center, and the number of gaze points that are closer to the center than the average distance to the center. In this case, the gaze center refers to the average point taken for all gaze points over the data segment considered. These features are shown in Fig. \ref{fig:gaze_features}. Before computing each gaze feature, the data was smoothed by running it through a five point moving average filter.

Several different features for the classifier were investigated. Although the ones selected are not independent, they provide a good estimate of the reliability of the predicted intent based on how focused the gaze is.

Training data for the classifier was gathered by recording all the eye data from the eye tracker while different volunteers looked at a screen filled with colored numbers. Each volunteer indicated their intent by clicking the space bar while looking at a number of their choosing. This caused the gazed-at-number to move to the center of the screen and labeled two seconds worth of the preceding valid data points with the class ``intent''. At anytime other time, gathered data was labeled with ``no intent''. Valid data includes all data where both eyes are fully tracked by the gaze tracker.

After the model has been trained, during actual intent prediction, each segment of data is taken from the last two seconds of valid collected gaze data and the gaze features are calculated. The classifier is then run on the input data.

The prediction output from the classifier includes the posterior probabilities of belonging to the classes ``intent'' and ``no intent''. These correspond to the operator beginning an action, or just observing the situation. Since there are only two classes, a posterior probability of over 50\% for ``intent'' indicates that it is most likely that the operator has a valid intention. However, a value of just over 50\% indicates that the intent is just barely likely. Therefore, the posterior probability for intent will be linearly rescaled to a range from 0.5 to 1.0:

\begin{gather} \label{eqn:iconf_scaling}
ci = 
\begin{cases} 
0 & p_i < 0.5 \\
\frac{1}{0.5} \left(p_i - 0.5\right) & p_i \geq 0.5 \\
\end{cases} \\
p_i = p\left(i|G_1,G_2,G_3\right) \nonumber
\end{gather}
where $ci$ is the confidence in the predicted intent and $p\left(i|G_1,G_2,G_3\right)$ is the probability of an intent given the three gaze features described. Additionally, if the operator's eyes are not tracked for over 0.75 seconds, the intent confidence is set to zero.

\section{Intent-based Haptic Assistance}

Two different haptic virtual fixtures are employed to provide assistance to the operator. Both are centered at the gaze-derived intent location. A guidance force pushes the operator's hand towards the target position with its strength based on distance from the target. Scaling the force this way ensures the system respects the operator's control. Even though the predicted intent location may be correct, the operator may need to navigate around obstacles so the guidance force should be relatively weak until the operator begins to move towards the target. Similarly, the safety boundary prevents destruction of the environment by preventing movement outside of a region close to the target.

\subsection{Haptic Guidance Force}

The purpose of the guidance force is to gently push the operator's hand towards the gaze-indicated target position. This will help overcome the lack of sensory feedback, especially in the depth direction. To this end, the profile of the guidance force is based on a method described in our previous work, termed potential hybrid control~\cite{potentialhybridcontrol}.

The potential hybrid controller takes two position inputs, a target (the gaze target, $\vec{p_g} \in \mathbb{R}^3$) and a manually-defined control position( joystick control location, $\vec{p_j} \in \mathbb{R}^3$). It then combines them in a way that takes into account the target uncertainty and the behavior of the operator. In particular, if the manually-defined control position is far, or very close, to the target position, then the potential hybrid controller does not affect the output very much. In the case of the manually-defined control position being far from the target, the operator should have complete control over the robot position. On the other hand, when the manually-defined control position is close to the target, the operator does not need much assistance because the target position has already been reached. It is only in the travel between these two extremes that assistance is needed. The potential hybrid controller follows this approach.

\subsubsection{Potential Hybrid Controller Method}
\label{sec:potential_hybrid_control}

\begin{figure}[htbp]
	\centering
	\includegraphics[width=0.75\linewidth]{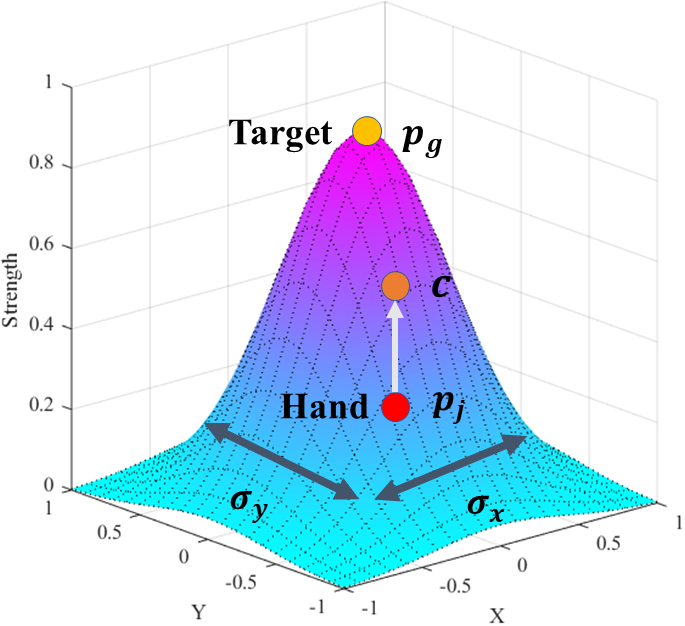}
	\caption{An illustration of the potential hybrid control method for a two-dimensional control space. $p_g$ is the goal point, $p_j$ is the joystick point, $c$ is the combined point, and $\sigma$ is the size of the field in each direction.}
	\label{fig:potential_method}
	%\vspace{-2cm}
\end{figure}
The target location, derived from the operator’s gaze, is combined with motion commands from the joystick through a potentially weighted influence method shown in Fig. \ref{fig:potential_method}. This approach uses the distance from $\vec{v_j}$ to $\vec{p_g}$ along a potential field to determine the influence of a resulting point, $\vec{c}$. The method is represented by the following equations:
\begin{equation}\label{eqn:combination}
\vec{c}=(\vec{p_j})(\vec{1.0}-W\vec{d})+(\vec{p_g} )W\vec{d}
\end{equation}
Where $\vec{c} \in \mathbb{R}^3$ is the final combined location. $W$ is the weight calculated from the potential field, and $\vec{d}$ is a 3x1 vector of coordinate weights,\begin{equation}
\vec{d} = \left[\begin{array}{c}d_x \\d_y \\d_z \end{array}\right]\nonumber
\end{equation}
From Eq. \ref{eqn:combination}, $W$ is the amount of influence that the target point $\vec{p_g}$ has on the resultant $\vec{c}$. It is calculated from the potential field and bounded by [0,1] where a higher value of $W$ approaching 1 means the resultant $\vec{c}$ will correspond to the target location, while a lower value of $W$ will mean it follows the hand position. $\vec{d}$ is a weighted vector that controls how the potential field affects the final combination in each direction. For example, a value of $d_x = 0.9$ would give the potential field a 90\% influence on the final combination for the x coordinate only.
\subsubsection{Potential Field}
A potential field describes how a body interacts with an entity that exerts an influence on that body. For example, in physics, there is a potential field description for the gravitational pull exerted by a planet. This potential field gives a representation of what forces another body would feel (due to the planet) when placed at any given location in the field. Analogously, the potential field in this method describes the effect the $\vec{p_g}$ has on the final position of the robot, $\vec{c}$.

The peak of the potential field is the gaze-indicated target position, this take advantage of the operator’s visuomotor behavior. By placing the potential field centered there, the robot's end-effector is drawn towards the intended target. The potential field provides a smooth combination or transition from the joystick position and the target position, which has its maximum effect at the intended location, however, this does not impact the robot position very much when the joystick is far from the target. This attribute of the potential field ensures the control follows the operator’s intent.As shown in Eq. \ref{eqn:potential_field}, a Gaussian curve was used for the potential field. While other potential fields could be used including parabolic, cubic, etc. ~\cite{gazecontingentmotor}, a Gaussian curve was selected because it is a smooth, continuous function and the shape is easy to manipulate by adjusting its parameters. The shape determines how quickly the influence of the field increases in each direction. 

\begin{gather}
\label{eqn:potential_field}
W = \exp \left(- \frac{1}{2} \left( \vec{p_j} - \vec{p_g} \right)^T \Sigma^{-1} \left( \vec{p_j} - \vec{p_g} \right) \right) \\
\Sigma^{-1} = \begin{bmatrix}
\sigma_{x}^2 & 0 & 0 \\
0 & \sigma_{y}^2 & 0 \\
0 & 0 & \sigma_{z}^2
\end{bmatrix} 
%\vspace{0.25cm}
\end{gather}
 
In the above equation $\vec{p_g}$ represents the center of the field and $\matr{\Sigma}$ represents the covariance matrix. The covariance matrix controls the tightness of the field. The smaller the variance in each directional component the tighter the field becomes. In this case, since the off-diagonal elements are zero so only the variance are left in the covariance matrix.  When this occurs, each coordinate direction in the field is independently controlled from one another.

There are a few reasons for using the method to combine the control inputs. One is the simplicity and ease of combining multiple inputs. Additionally, the method is intuitive from a physical sense. The potential field represents the probability the target location determined from the gaze is correct. As the operator joystick command motion approaches the target position, the system becomes more confident of its own guess at the target location and continues to increase its own influence over the robot end-effector. If, on the other hand, the robot end-effector is far away from the gaze-selected target then the system has a lower confidence in the target location so it affects the final end-effector location less. This also models the way humans naturally behave. If one is looking at a particular location in space, they do not want their hand moved there automatically, but if they are focusing on something intently (for example when threading a needle) their hand moves to wherever they are looking.

\subsubsection{Potential Hybrid Control Approach Guidance Force}
After determining the location of $\vec{c}$ in the previous section, the guidance force for the haptic feedback must be determined. By basing the strength of the guidance force on the potential hybrid control, all the advantages of the method noted above, are gained. The strength of this force is proportional to the degree of influence given by the potential hybrid control for a given direction. It is calculated using:
\begin{equation}
\vec{gf} = \frac{\vec{c} - \vec{p_j}}{gf_{max}}
\end{equation}
where $\vec{gf} \in \mathbb{R}^3$ is the normalized, dimensionless strength of the guidance force in Cartesian space, $\vec{c}$ is found using (\ref{eqn:combination}), $\vec{p_j}$ is the normalized joystick position, and $gf_{max}$ is the maximum value of $\vec{c} - \vec{p_j}$. $gf_{max}$ is pre-calculated using numerical methods. $gf$ is then scaled to an appropriate level for the haptic device to give the actual guidance force. The scaling ensures the joystick will not pull itself out of a user's hand. A size of $\sigma = 0.4$ was used for all coordinate directions, as suggested in our previous work. In the simplest case, the direction of the force points directly towards the target. However, just like the amount of influence for each direction could be controlled with the potential hybrid control approach described in section \ref{sec:potential_hybrid_control}, the strength of the force in each direction can be controlled independently. In most cases, the strength in the depth direction should be much larger than the other directions since that is the direction lacking feedback for the operator.

\begin{figure}[!htbp]
	\centering
	\includegraphics[width=0.9\linewidth]{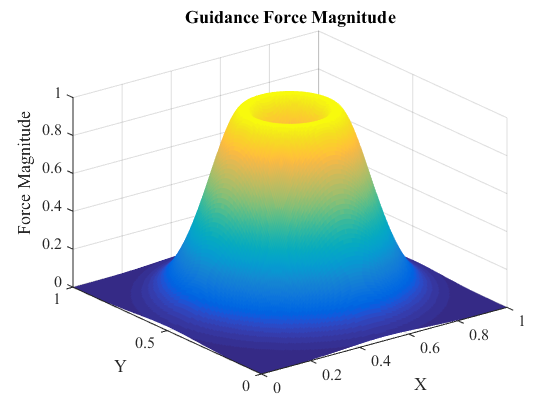}
	\caption{Magnitude of the guidance force in two dimensions for a combination with equal strengths in both $X$ and $Y$. The target position is placed at (0.5, 0.5). The profile of the force is the same for all directions of approach.}
	\label{fig:force_mag}
	%\vspace{-0.5cm}
\end{figure}

\begin{figure}[!htbp]
	\centering
	\includegraphics[width=0.7\linewidth]{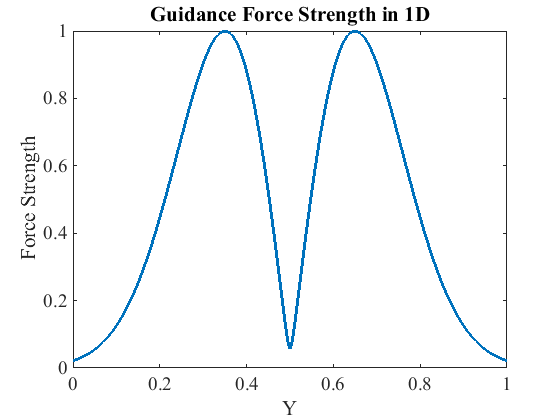}
	\caption{Magnitude of the guidance force in one dimension. The target position is placed at $Y = 0.5$.}
	\label{fig:force_1d}
	%\vspace{-0.5cm}
\end{figure}

The effect of the guidance force is to provide a gentle push in the direction of the target when the joystick is moving towards the target. Similar to the potential hybrid control influence, if the joystick is far from the target, or close to the target, then the force is small, but otherwise the force is larger. An illustration of the magnitude of the force (strength in each direction combined together) is presented in Fig. \ref{fig:force_mag} in two dimensions for the case where the strengths, $d_X$ and $d_Y$, in each direction are equal. The force profile is the same from every direction of approach. This is also illustrated in Fig. \ref{fig:force_1d}, which shows the force magnitude for a single direction, or can be thought of as a section view of Fig. \ref{fig:force_mag}.

\begin{figure}[!htbp]
	\centering
	\includegraphics[width=0.7\linewidth]{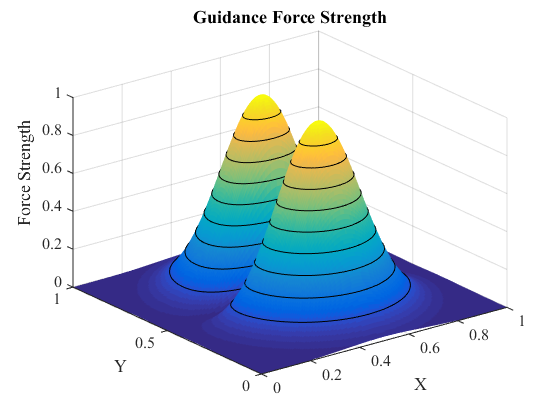}
	\caption{Strength of the guidance force in two dimensions for a combination where only the depth is applied. $Y$ is the depth direction in this case and the only direction with force assistance. The target position is placed at (0.5, 0.5).}
	\label{fig:force_depth}
%	\vspace{-0.5cm}
\end{figure}

Figure \ref{fig:force_depth} shows the normalized strength when the force is only applied in the depth direction, $Y$. Moving along the depth direction at the point where $X$ is equal to the target position, $X = 0.5$, gives the same profile as the magnitude plot in Fig. \ref{fig:force_mag}. This is helpful to the operator, because the depth direction is the only direction in this situation that requires assistance. At other points, at values of $X$ increasingly farther from the target, the peak strength of the guidance force decreases. This allows the operator to control the robot with less opposition if the target position is incorrect, or the approach path needs to deviate from its current trajectory, for example due to an obstacle. Along the line $Y = 0.5$ the force is completely zero. This is because the target position in the depth direction has already been reached, so minimal additional assistance is necessary.

\subsection{Safety Boundary}
\label{sec:safety_boundary}

An alternative to the guidance force, a haptic safety boundary is considered. The purpose of the safety boundary (forbidden-region virtual fixture) is to minimize collateral damage by restricting movement of the joystick to a small area surrounding $\vec{p_g}$ Various different shapes for this boundary are examined in the following sections. The shape of the boundary is very important because the boundary has the potential to make teleoperation safer, but it could also prevent the operator from controlling the joystick effectively if the shape is not chosen carefully, which would lead to more mistakes and environmental damage.

\subsubsection{Boundary Design}

\begin{figure}[!htbp]
	\centering
	\includegraphics[width=0.6\linewidth]{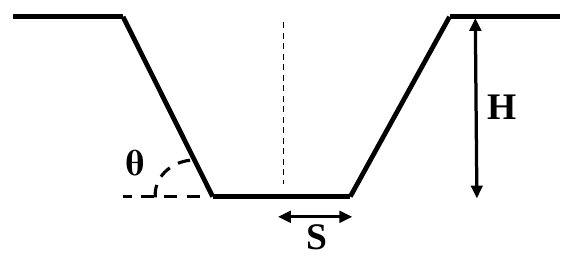}
	\caption{Safety boundary shape with parameters.}
	\label{fig:boundary_shape}
%	\vspace{-0.5cm}
\end{figure}

The shape of the safety boundary, shown as a section-view in Fig. \ref{fig:boundary_shape}, was chosen to provide minimum intrusion to the operator's standard operating manner while still restricting access to areas unnecessary for completing the task. The full shape can be created by rotating the profile in Fig. \ref{fig:boundary_shape} by pi radians about its center axis. The upper plane prevents unintended damage during general motion, while the cone allows the robot room to move in and complete the task at $\vec{p_g}$.

There are three parameters that govern the shape of the boundary: $S$, $H$, and $\theta$. $S$ is the radius of the flat bottom, $H$ is the height limit of the cone, and $\theta$ is the angle of the cone. In general, a small $S$, and a large $H$ and $\theta$ will create the most restrictive boundary. For the purposes of this system, $S$ was limited from 1 centimeter to 7 centimeters, $H$ was limited from 0 to 15 centimeters, and $\theta$ from $5^{\circ}$ to $85^{\circ}$. These limits were chosen based on the physical limitations of the haptic device and on empirical experience of the necessary room needed to maneuver to complete a task.

\subsubsection{Parameter Selection}

Selecting the correct set of parameters for the safety boundary is important because of the competing aims of adjustment. Tightening the safety boundary, which corresponds to increasing $H$, decreasing $S$, and increasing $\theta$ can reduce the risk of collateral damage by preventing access to areas of the workspace farther from $\vec{p_g}$. However, while this may reduce the risk of damage, it increases the risk of failure. With such a tight boundary, there may not be enough room to maneuver. At the extreme, if $\theta = 90^{\circ}$, $S = 0$, and $H > 0$ then the workspace is not accessible at all. Opening up the boundary, decreasing $H$, increasing $S$, and decreasing $\theta$ can lead to the opposite problem: high risk of collateral damage, but a high likelihood of success.

The set of parameters for each task should be chosen to minimize the overall risk associated with the task, including the risk of damage and the risk of failure. There is no clear set of parameters to achieve this, and the selection is task dependent. Therefore, it is necessary to investigate how the variation of parameters affects the risks. This was done by measuring the failure rate of each task while using different boundary parameters. Failure rate considered both failures due to not completing the task, and failures due to damaging the surroundings. The set of parameters with the minimum failure rate was selected as the initial set for each task. Two different tasks were measured, grasping and cutting. These are explained in more detail in section \ref{sec:haptic_validation_process}.

Eight different combinations of the parameters were tested. The combinations tested were selected by dividing the range of each parameter into thirds and using the cutoffs between each segment. This was done to test each parameter at a high and low point, and also to leave room for adjustment later. The specific combinations of parameters measured for each task are shown in Table \ref{tab:boundary_sets}. A mouse click was used to set the center of the boundary on the target position for each task.

\begin{table}[!htbp]
	\caption{\textsc{Safety Boundary Parameter Sets}}
	\label{tab:boundary_sets}
	\renewcommand{\arraystretch}{1.3}
	\centering
	\begin{tabular}{|c|c|c|c|}
		\hline
		\textbf{Set} & $\mathbf{\theta~[deg]}$ & $\mathbf{H~[cm]}$ & $\mathbf{S~[cm]}$\\
		\hline
		1 & 30 & 5 & 3\\
		\hline
		2 & 30 & 5 & 5\\
		\hline
		3 &	30 & 10 & 3\\
		\hline
		4 &	30 & 10 & 5\\
		\hline
		5 &	60 & 5 & 3\\
		\hline
		6 & 60 & 5 & 5\\
		\hline
		7 & 60 & 10 & 3\\
		\hline
		8 & 60 & 10 & 5\\
		\hline
	\end{tabular}
\end{table}

Eight volunteers were tested, with each one performing each task under every set described in Table \ref{tab:boundary_sets} for two to three trials. All the volunteers were aged 18-28 and were able-bodied. Three of the volunteers had prior experience using the system, but all were given as much time as they needed to familiarize themselves with how it worked. This was done by practicing picking up a tennis ball with no boundary until they felt comfortable with the system. The order of the sets was randomized for every participant to prevent acclimation to the system, which would affect the results. The failure rate was then averaged for each trial and over all the participants to determine the overall failure rate for every parameter set. 

The results of the tests are shown in Table \ref{tab:boundary_param_results}. As expected, the average failure rate for the cutting task, 59\%, was higher than for the grasping task, 54\%. This is simply because the cutting task is more difficult and requires higher precision while manipulating the robot arm. For the cutting task, there are multiple sets which have equal failure rates; sets 2, 6, and 7 have the lowest failure rate of 50\%. These sets correspond to the least restrictive boundary (set 2) and the most restrictive boundary (set 7), as well as a less restrictive boundary (set 6). This could be explained by the difficulty of the task. Since the task was challenging, it was easiest to accomplish with low restrictions, which would give the most room to maneuver, or with high restrictions which would provide the most protection from collateral damage. Further testing would likely help to differentiate between these sets. Any one of these sets can be chosen to minimize the risk. Since the cutting task is difficult, set 2 was chosen to proceed with because it will give the operator more control. For the grasping task, set 5 had the minimum failure rate.

\begin{table}[!htbp]
	\caption{\textsc{Safety Boundary Parameter Test Results Showing the Failure Rate of Each Task. The Highlighted Cell(s) in Each Column Indicate the Parameter Set(s) with the Minimum Failure Rate}}
	\label{tab:boundary_param_results}
	\renewcommand{\arraystretch}{1.3}
	\centering
	\begin{tabular}{|c|c|c|}
		\hline
		\textbf{Set} & \textbf{Cutting Task} & \textbf{Grasping Task}\\
		\hline
		1 & 64\% & 65\%\\
		\hline
		2 & \cellcolor{blue!15} 50\% & 42\%\\
		\hline
		3 &	64\% & 58\%\\
		\hline
		4 &	64\% & 73\%\\
		\hline
		5 &	71\% & \cellcolor{blue!15} 38\%\\
		\hline
		6 & \cellcolor{blue!15} 50\% & 65\%\\
		\hline
		7 & \cellcolor{blue!15} 50\% & 48\%\\
		\hline
		8 & 57\% & 44\%\\
		\hline
	\end{tabular}
\end{table}

\section{Adjustment Based on Intent Confidence}
\label{sec:confidence_adjustment}

Both the guidance force and the safety boundary are adjusted from their initial settings based on the level of confidence in the operator's intent. Specifically, the safety boundary will become less restrictive and the guidance force weaker when the intent confidence is low while the opposite will occur when the intent confidence is high. The reason for this is that the operator should not be restricted when the system is not confident in its prediction of the intent. This would cause frustration, errors, and possible damage to the surrounding environment as the operator has to fight the system to get to where she actually wants to go. On the other hand, if the system is highly confident in the predicted intent, then the strength of the haptic assistance should be increased to guide the operator to the target position and minimize the risk of damaging the environment.

Therefore, the safety boundary will open up when the confidence level is low, and tighten when the confidence level is high. Specifically, $S$ and $\theta$ will be increased while $H$ is decreased when the confidence is low. This relationship is shown in (\ref{eqn:confidence_adjustment}).

\begin{subequations} \label{eqn:confidence_adjustment}
	\begin{gather} 
	sci = 
	\begin{cases} 
	\frac{ci-ithresh}{1-ithresh} & ci \geq ithresh \\
	\frac{ci-ithresh}{ithresh} & ci < ithresh \\
	\end{cases} \\
	paramAdjust = sci*paramMaxAdjustAmount
	\end{gather}
\end{subequations}
where $sci$ is the scaled confidence in the intent, $ci$ is the confidence in the predicted intent calculated using (\ref{eqn:iconf_scaling}), $ithresh$ is the confidence threshold level, $paramAdjust$ is the amount to adjust one of the parameters, and $paramMaxAdjustAmount$ is the maximum amount that the parameter can be adjusted. The confidence threshold level is the confidence level at which the intent prediction is high enough to begin to make the safety boundary more restrictive. It is suggested that this be set to a value over 50\% because that is the point at which the system is more confident in the intent than not. For this method, $ithresh$ was set to 60\%. Once the scaled confidence, $sci$, is computed, each parameter is adjusted from its initial value based on the $paramMaxAdjustAmount$. The value of this parameter for $S$, $H$, and $\theta$ was chosen as a third of the range for each parameter, as described in section \ref{sec:safety_boundary}. In particular, $SMaxAdjustAmount = -2~\text{cm}$, $HMaxAdjustAmount = 5~\text{cm}$, and $\theta MaxAdjustAmount = 25^{\circ}$. This will allow each parameter to be scaled to the maximum or minimum of its range depending on the intent confidence. For example, $S$ will scale down to its lowest value of 3 centimeters when the intent confidence is 100\%. This process enables the safety boundary to dynamically adjust to the confidence level in order to decrease the risk of damaging the environment.

\begin{figure*}[!htbp]
	\centering
	\begin{subfigure}{\columnwidth}
		\centering
		\includegraphics[width=0.85\columnwidth]{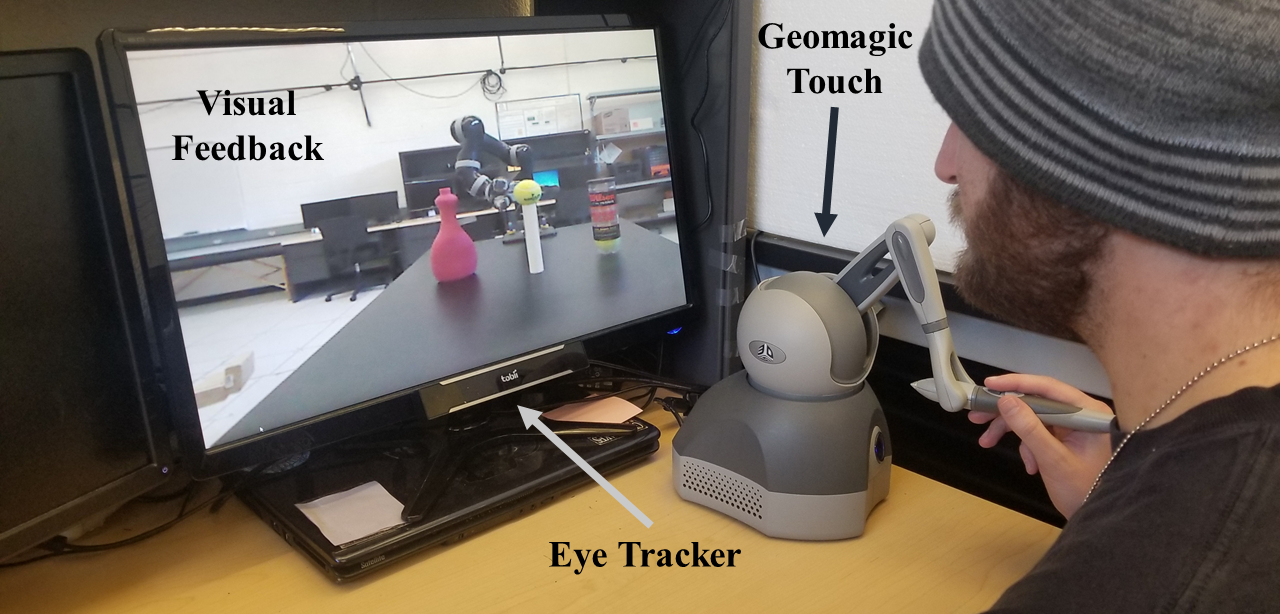}
		\caption{Operator-side setup}
		\label{fig:experiment_setup_first}
	\end{subfigure}\hfill
	\begin{subfigure}{\columnwidth}
		\centering
		\includegraphics[width=0.85\columnwidth]{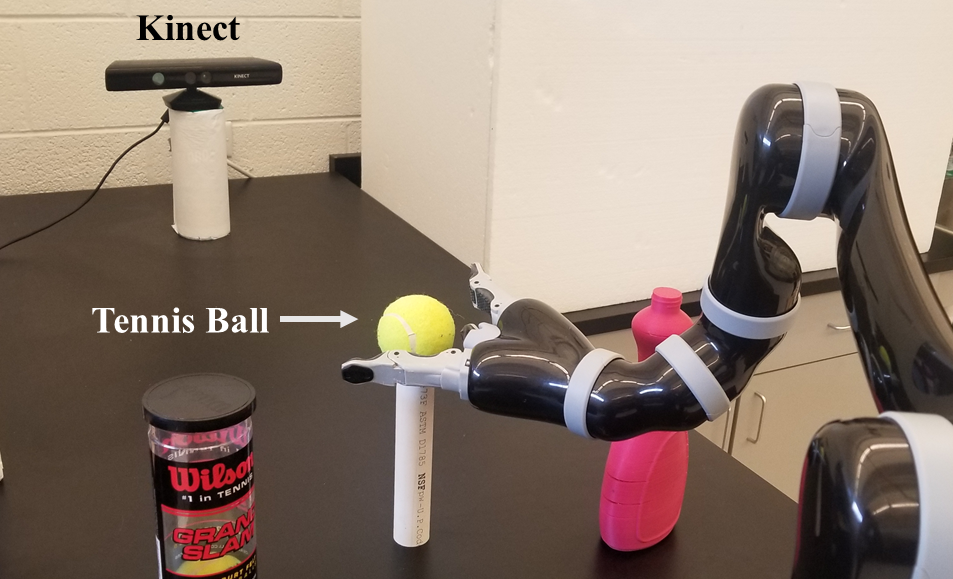}
		\caption{Robot-side setup}
		\label{fig:experiment_setup_second}
	\end{subfigure}
	\caption{Experimental setup which shows both the operator-side and the robot-side setup.}
	\label{fig:experiment_setup}
%	\vspace{-0.5cm}
\end{figure*}

Additionally, the guidance force strength is also adjusted based on the confidence level. In this case, it is applied as a simple linear scaling, meaning that if the scaled confidence level is 0 then no guidance force is applied to the operator, and if the scaled confidence level is 1, then the full strength of the guidance force is exerted on the operator.

\section{Experimental Validation}
\label{sec:experiment}

In order to evaluate the effectiveness of the proposed approach, each component was tested with the setup described in the following section. Both the guidance force and the safety boundary were tested separately to gain an understanding of how each affected the teleoperation performance.

\subsection{Experiment Setup}

For validation, a system was built following Fig. \ref{fig:fig_overall}. The joystick used is a Geomagic Touch created by 3D Systems. This is a haptic device with 6 degrees of freedom that was configured to output the pose in space represented by the stylus. Additionally, a small amount of constant friction was applied to stabilize motion of the stylus and make it easier for the operator to produce precise adjustments. The eye-tracking portion of the project was based on the Tobii Rex eye tracker. This eye tracker is a video-based remote system which can track the user's eyes from 40-90 cm away and allows significant head movement as long as it stays inside the trackable volume. To determine the fully specified target position, a structured light sensor, the Microsoft Kinect, was used. The Kinect provides a depth image at a resolution of 640x480 that is used to determine the depth of the target position. The Kinect also supplies the video feedback looking straight-on to the scene. The robot arm used was a three-fingered Mico robot from Kinova. Opening and closing of the robot's fingers was controlled by a button on the Geomagic Touch. Updates to change the desired position or state of the fingers were sent to the robot at approximately 20 Hz, unless the joystick was immobile and the fingers were not being controlled. The experimental setup can be seen in Fig. \ref{fig:experiment_setup}.

\subsection{Validation Process}
\label{sec:haptic_validation_process}

\begin{figure*}[!bhtp]
	\centering
	\begin{subfigure}{\columnwidth}
		\centering
		\includegraphics[width=0.8\columnwidth]{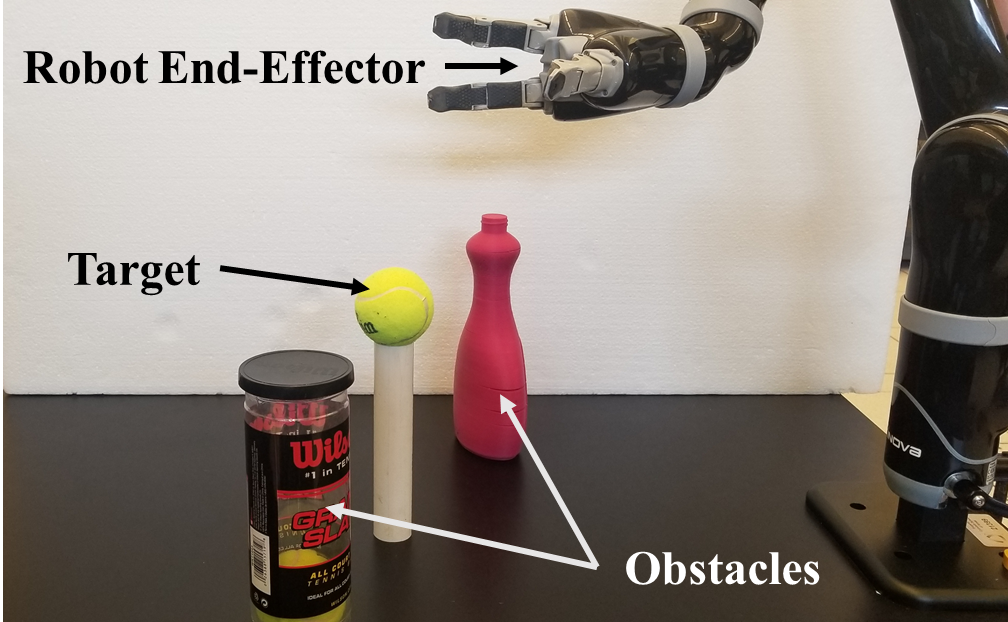}
		\caption{Grasping task setup.}
		\label{fig:grasping_task}
	\end{subfigure}\hfill
	\begin{subfigure}{\columnwidth}
		\centering
		\includegraphics[width=0.6\columnwidth]{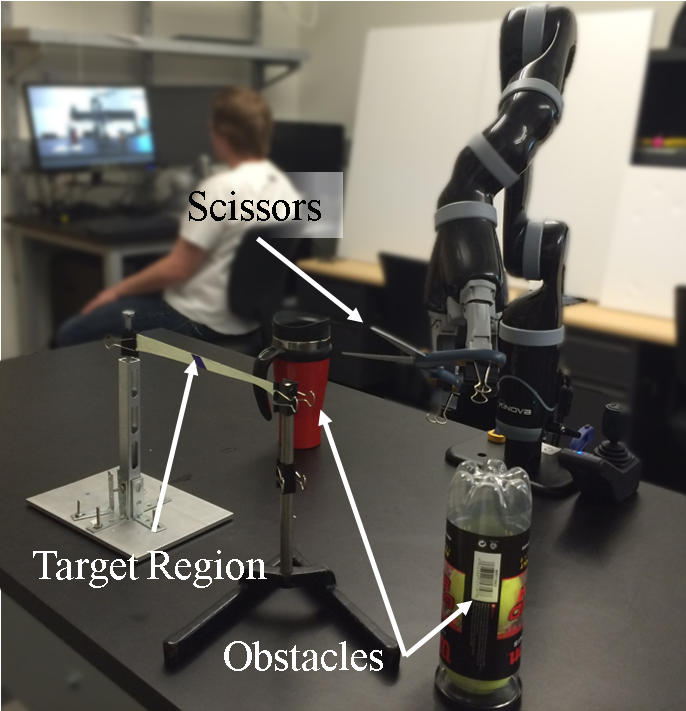}
		\caption{Cutting task setup.}
		\label{fig:cutting_task}
	\end{subfigure}
	\caption{Task setups used for testing the haptic assistance.}
	\label{fig:task_setup}
	%\vspace{-0.5cm}
\end{figure*}

Two different tasks were tested in the experiment, cutting and grasping. For the grasping task, volunteers were asked to use the teleoperation system to pick up a tennis ball. The task was considered a failure if the tennis ball was knocked off its stand, or if any of the surrounding obstacles were disturbed. The setup for this task is shown in Fig. \ref{fig:grasping_task}. For the cutting task, volunteers were asked to cut a strip of paper in a special marked area using the teleoperation system. This is illustrated in Fig. \ref{fig:cutting_task}. Failure occurred during this task if the strip was cut in the wrong location or the surrounding area was harmed. This task simulated an action similar to one that might be required in telesurgery.

The testing procedure began by calibrating the eye tracker for each volunteer and verifying its accuracy. Each volunteer was then given as much time as they needed to become comfortable with the system, or re-familiarize themselves with it if they had already used it. No force feedback was applied during this part, and practice was done on the grasping task with no obstacles. Once they were ready, each task was tested with both the guidance force and the safety boundary in a randomized order to ensure results were not skewed by a learning curve. Additionally, both tasks were tested without any haptic assistance (using the joystick only) to provide a baseline for comparison. Explicitly, the system was tested in the combinations laid out in Table \ref{tab:haptic_test_sets}. Two to three trials were performed for each combination and the success rate, as well as the joystick and robot trajectory were recorded. The target, either the tennis ball or paper strip, was randomly placed for each trial. The target position was continuously acquired from the gaze and before each trial the Geomagic Touch was placed into a starting position that the robot mirrored.

Four volunteers were tested. The ages of those who participated in the testing were in the range 18 to 25 and two of the volunteers wore glasses. One was left-handed and three had prior experience using the system.

\begin{table}[!htbp]
	\caption{Combinations tested for the validation of the haptic assistance.}
	\label{tab:haptic_test_sets}
	\renewcommand{\arraystretch}{1.3}
	\centering
	\begin{tabular}{|c|c|c|}
		\hline
		\textbf{Test} & \textbf{Task} & \textbf{Haptic Assistance}\\
		\hline
		1 & cutting & safety boundary \\
		\hline
		2 & cutting & guidance force \\
		\hline
		3 &	grasping & safety boundary \\
		\hline
		4 &	grasping & guidance force \\
		\hline
		5 & cutting & no assistance \\
		\hline
		6 & grasping & no assistance \\
		\hline
	\end{tabular}
\end{table}

\section{Results and Discussion}
\label{sec:experimental_results}

During each trial, the target location, indicated by the gaze, and the trajectory of the Geomagic Touch and the robot arm was recorded. Additional information recorded for each trial included the number of times the participant tried to grasp the ball by closing the fingers and whether or not they were finally successful. The results are broken down into two separate tasks, cutting and grasping. For each we will evaluate three criteria: success rate, completion time, and attempts by the operator to close the scissors/fingers. 

\subsection{Cutting Task}

\subsubsection{Success Rate}
\begin{figure}[!htbp]
	\centering
	\includegraphics[width=\linewidth]{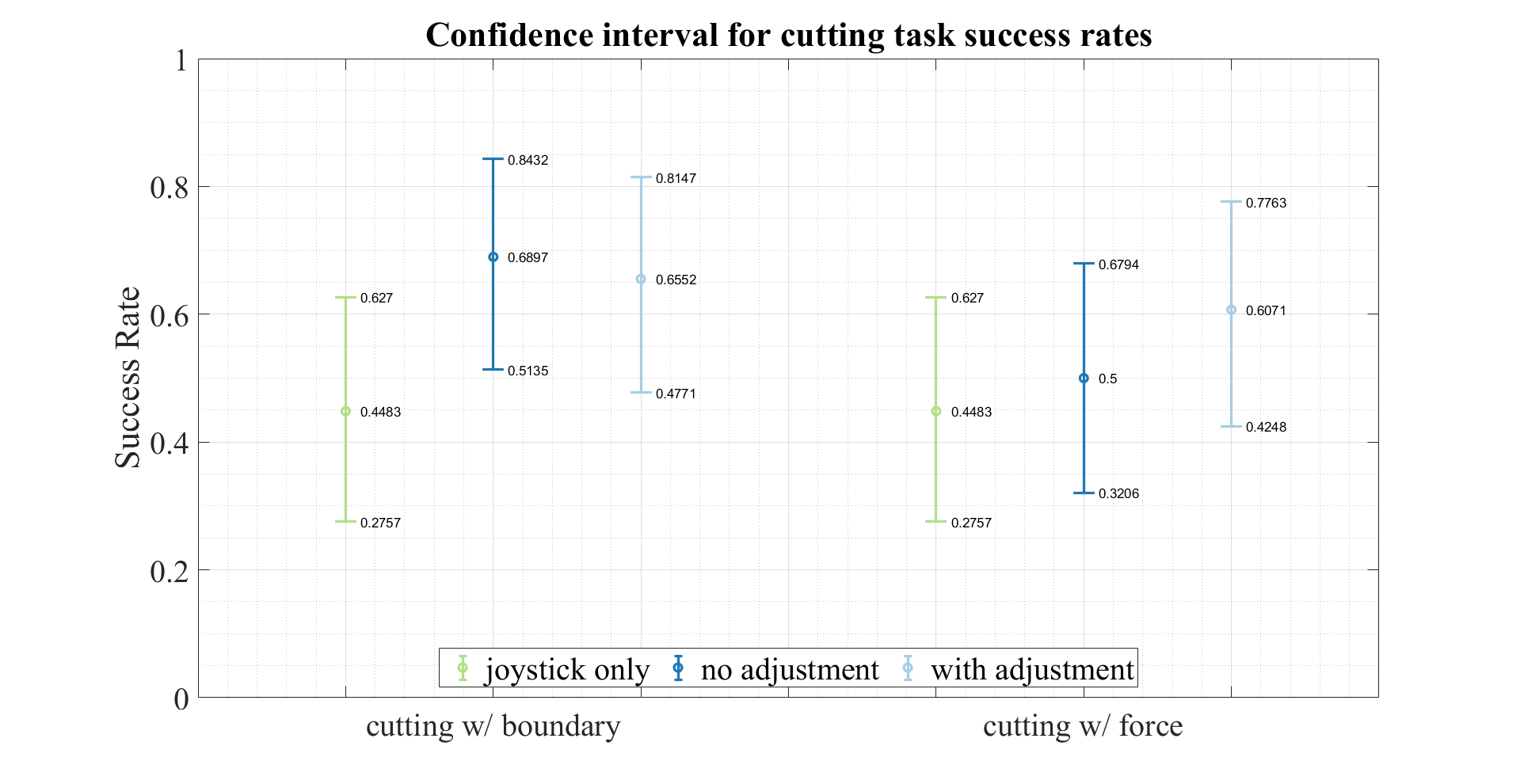}
	\caption{Confidence intervals for the success rates for each combination of haptic assistance per task.}
	\label{fig:cut_success}
	%	\vspace{-0.5cm}
\end{figure}

Due to the number of trials obtained, a Laplace estimate is used to determine the best success rate for each condition. Further a 95\% adjusted-Wald Interval is used to compare the theoretical bounds on the success rate observed. The confidence intervals with the Laplace estimate are observed in Figure \ref{fig:cut_success}.  For ease of reading, the joystick success rate has been duplicated on the figure for both control modes. In both cases, the assistance improves the success rate over the joystick only control. The boundary assistance outperforms the force guidance assistance. The intent adjustment appears to help the force guidance mode perform better (likely due to impacting the magnitude of the force more intuitively). Yet, the intent adjustment in the boundary assistance does not see improvement, although the success rates are rather similar. This is likely due to the users perceiving this change as subtle adaption. An N-1 Chi-Square test was conducted to determine statistical significance between any proportions. No statistical significance was found. 

\subsubsection{Execution Time}
\begin{figure}[!htbp]
	\centering
	\includegraphics[width=\linewidth]{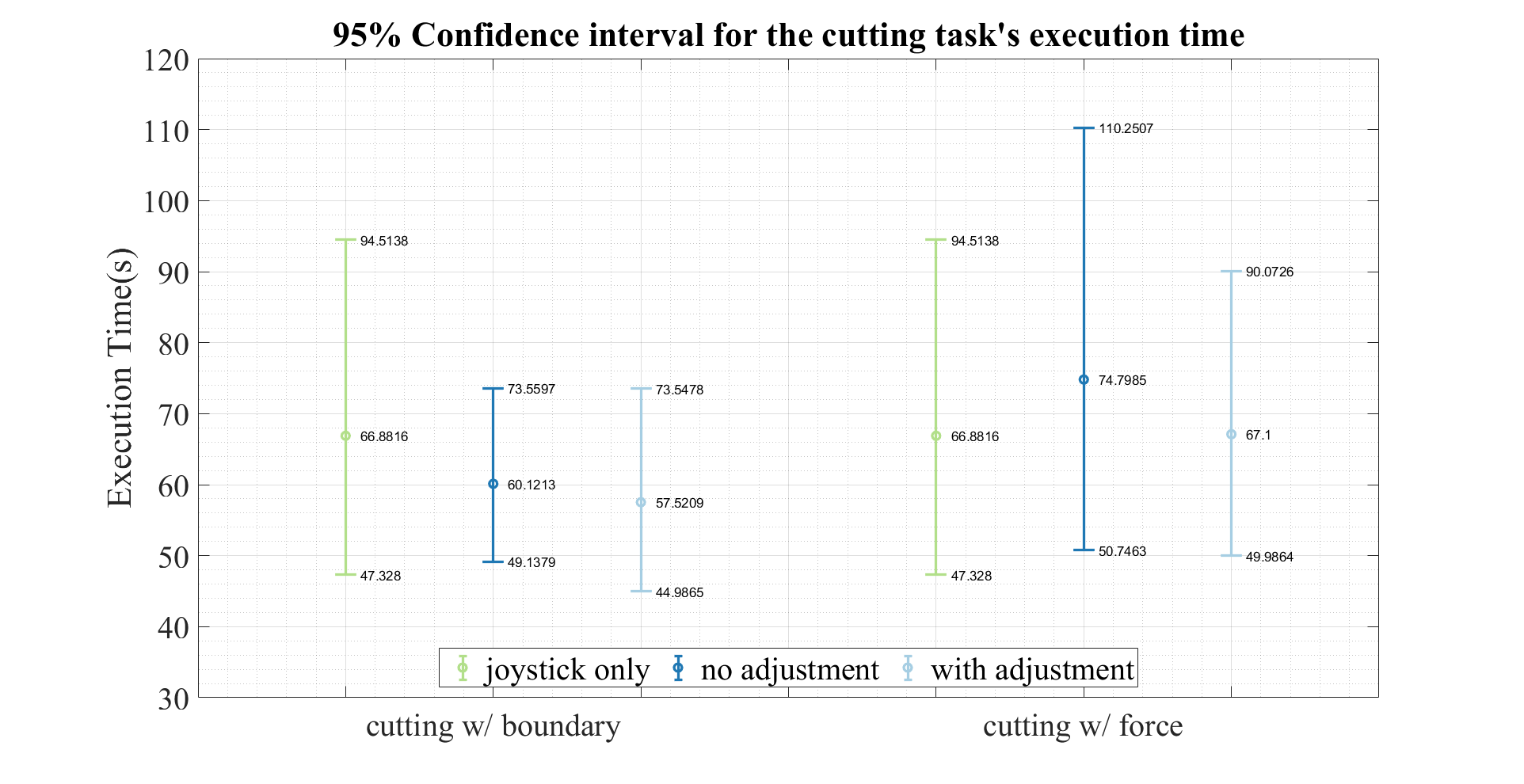}
	\caption{Confidence intervals for execution time for each combination of haptic assistance per task.}
	\label{fig:cut_time}
	%	\vspace{-0.5cm}
\end{figure}

Time based evaluations are notorious for being positively skewed \cite{10.1145/1753326.1753679}, and for this reason the analysis is done by log-transforming the data. The geometric means and 95\% confidence intervals are presented in Figure \ref{fig:cut_time}. The boundary assistance does better than joystick control. The boundary assistance also outperforms the force guidance.  The intent adjustment helps both the boundary and guidance force improve the speed to complete the task. However, the improvement is more noticeable in the force guidance. A two-sample t-test was conducted on each condition, and no statistical significance was found. Although no significance was found, the confidence intervals reinforce that the boundary approach is a better control strategy. It has the smallest bounds while accomplishing the least amount of time. The force guidance appears to be a hindrance to users as if it requires effort to resist undesired movements. However, the extra time observed from the guidance force may be a result incorrect depth registering as evident by the cutting attempts.

\subsubsection{Cutting Attempts}
\begin{figure}[!htbp]
	\centering
	\includegraphics[width=\linewidth]{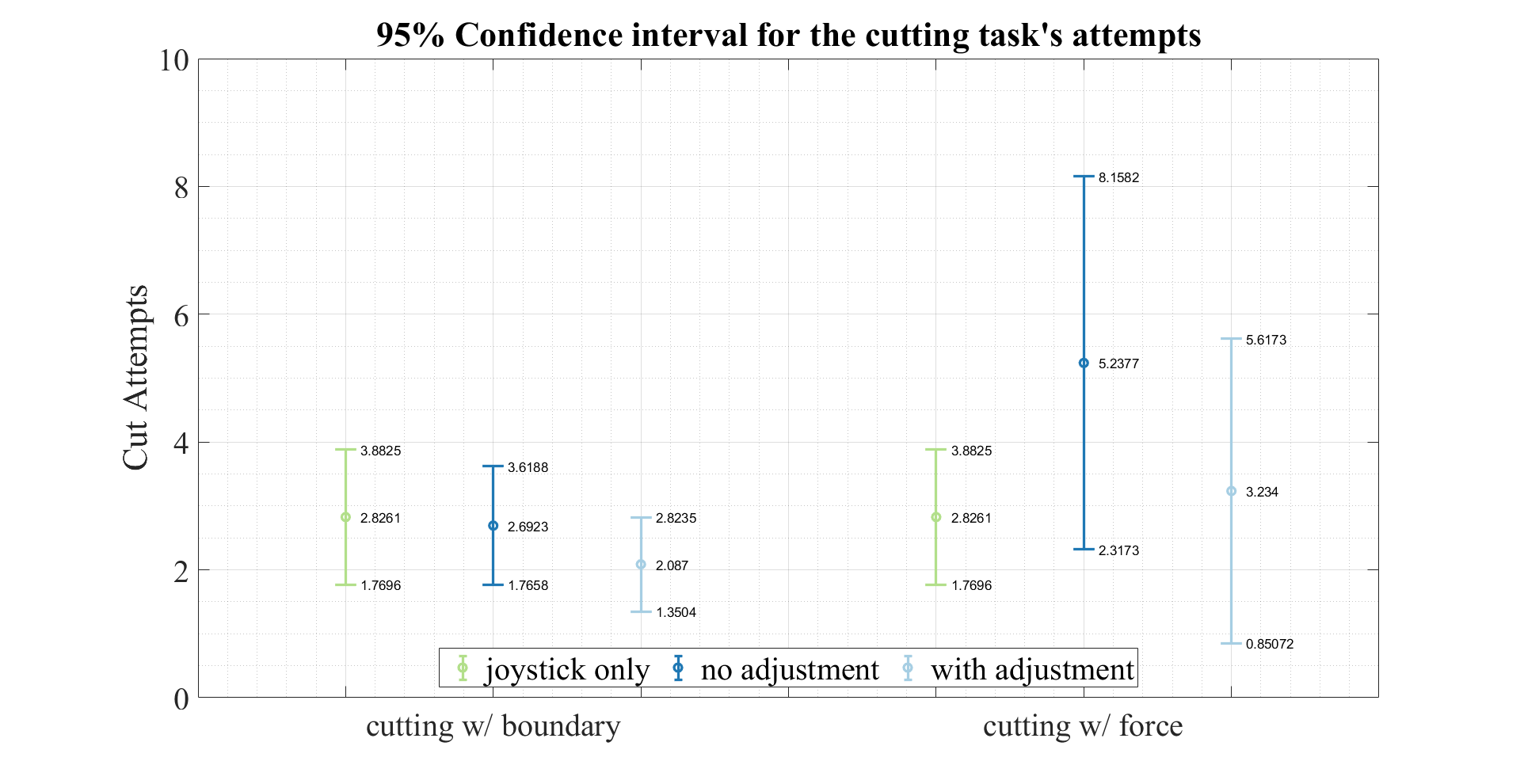}
	\caption{Confidence intervals for the cutting attempts for each combination of haptic assistance per task.}
	\label{fig:cut_attempts}
	%	\vspace{-0.5cm}
\end{figure}

A standard arithmetic mean and 95\% confidence interval was obtained for the cutting attempts of each control strategy. They are displayed in Figure \ref{fig:cut_attempts}. The confidence intervals of the boundary assistance are as low or lower than the joystick control. The guidance force required users to make more attempts to cut. This surge in attempts is most likely responsible for the time increase. The intent adjustment reduces the number of attempts needed to accomplish the task. A two-sample t-test was conducted for intent adjusted vs not intent adjusted control modes for a fair comparison. No statistical significance was found. 

\subsection{Grasping Task}

For the grasping task, only the assistance modes are compared. The goal of this task is to see if the intent adjustment is different from no intent adjustment. Issues from this task occurred when the robot hand would occasionally be the inferred gaze target.

\subsubsection{Success Rate}

\begin{figure}[!htbp]
	\centering
	\includegraphics[width=\linewidth]{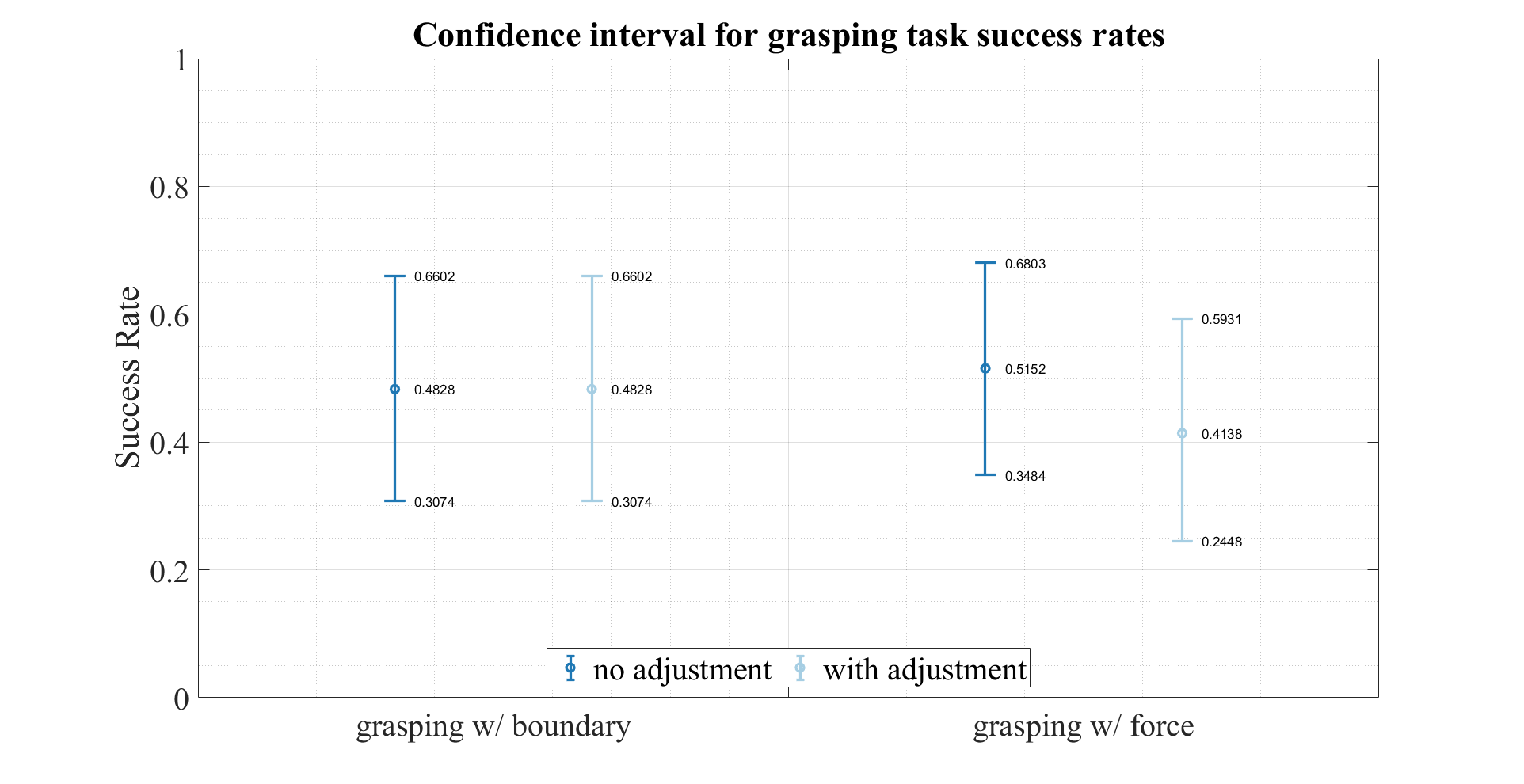}
	\caption{Confidence intervals for the grasping success for each combination of haptic assistance per task.}
	\label{fig:grasp_success}
	%	\vspace{-0.5cm}
\end{figure}

Due to the number of trials obtained, a Laplace estimate is used to determine the best success rate for each condition. Further a 95\% adjusted-Wald Interval is used to compare the theoretical bounds on the success rate observed. The confidence intervals with the Laplace estimate are observed in Figure \ref{fig:grasp_success}. The intent adjustment does not have a positive influence on the success rate. In the force guidance case, it appears to lower the success. An N-1 Chi-Square test was conducted to determine statistical significance between any proportions. No statistical significance was found.

\subsubsection{Execution Time}
\begin{figure}[!htbp]
	\centering
	\includegraphics[width=\linewidth]{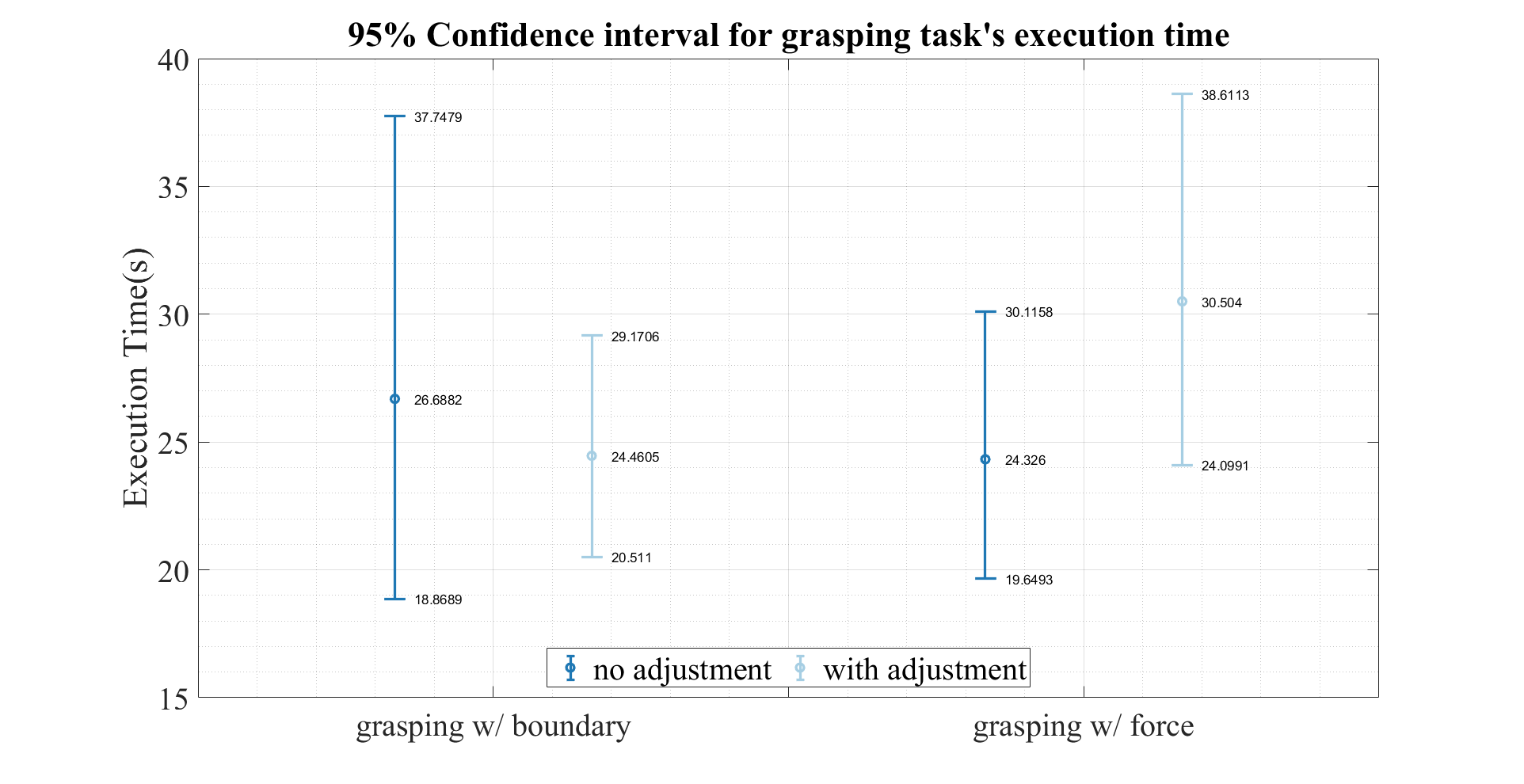}
	\caption{Confidence intervals for the grasping execution time for each combination of haptic assistance per task.}
	\label{fig:grasp_time}
	%	\vspace{-0.5cm}
\end{figure}

The geometric means and 95\% confidence intervals are presented in Figure \ref{fig:grasp_time}.  The completion time for the grasping task leads to mixed results. The intent adjustment helps the boundary approach; however, it does not help the guidance force assistance. For the no intent adjustment cases, the force guidance does better than the boundary. For intent adjustment, the boundary outperforms the guidance force. A two-sample t-test was conducted on each condition. Despite the variations of the confidence intervals no statistical significance was found. 

\subsubsection{Grasping Attempts}

\begin{figure}[!htbp]
	\centering
	\includegraphics[width=\linewidth]{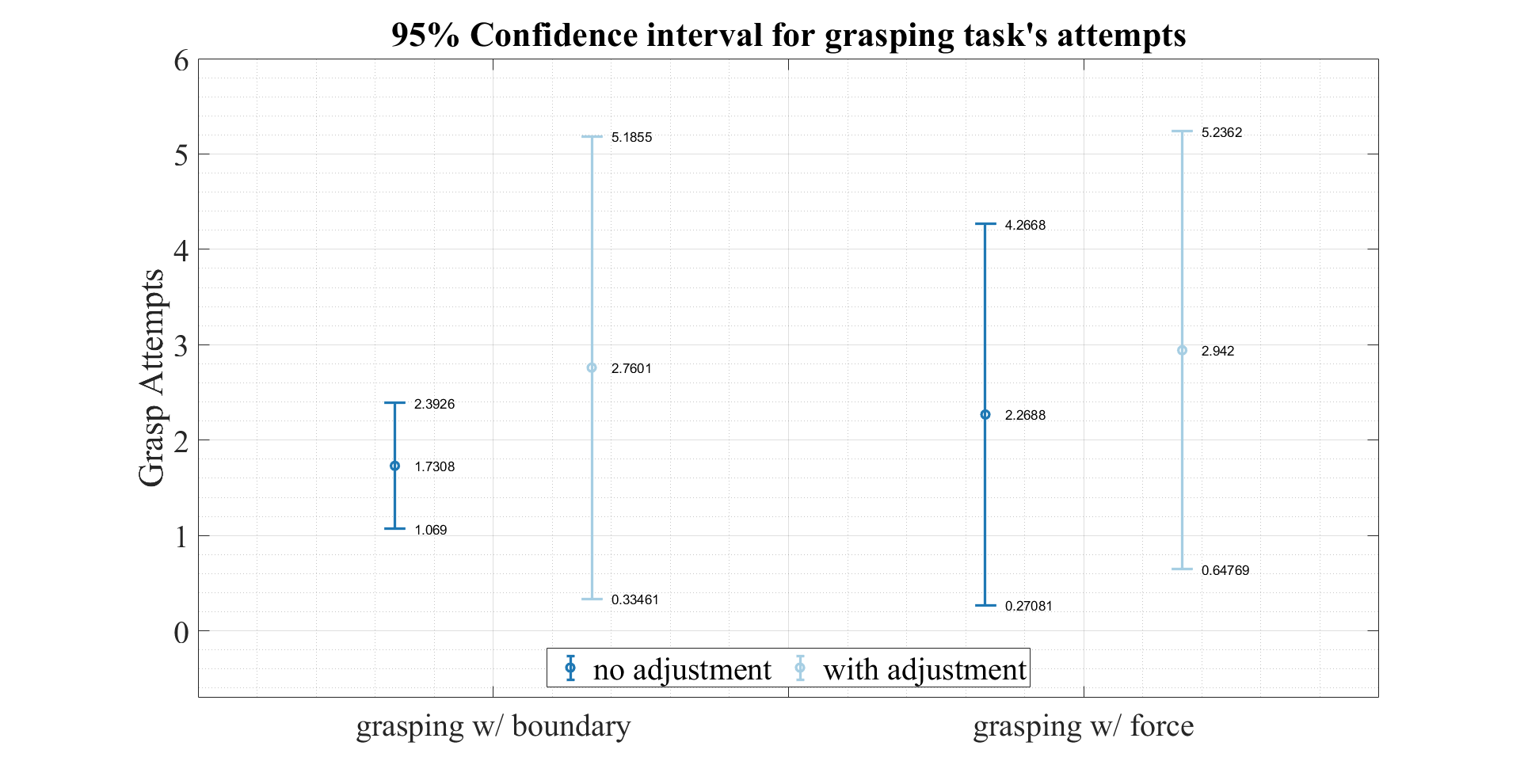}
	\caption{Confidence intervals for the grasping attempts for each combination of haptic assistance per task.}
	\label{fig:grasp_attempt}
	%	\vspace{-0.5cm}
\end{figure}

A standard arithmetic mean and 95\% confidence interval was obtained for the grasping attempts of each control strategy. They are displayed in Figure \ref{fig:grasp_attempt}. The intent adjustment forces more attempts to occur to grasp the tennis ball. The best scenario is the boundary without intent adjustment. A two-sample t-test was conducted for adjusted vs not adjusted control modes for a fair comparison. No statistical significance was found. 

\subsection{Summary of Results}

In summary, the presented results show that this system improves teleoperation control by assisting the operator in reaching the correct target depth and preventing collateral damage. In addition, the intent confidence is a valuable addition to the approach which allows the system to respond to the operator's focus, and provides more natural control for the operator.

The results also reveal some details about how each form of haptic assistance affects each task. While the safety boundary with intent confidence appears to be quite helpful, the same is not true of the guidance force. For maximum success, the intent confidence adjustment should not be used with the guidance force on the grasping task as this addition decreased the success rate.  Furthermore, it seems that the guidance force for the cutting task is not as helpful. It may be a better idea to use the safety boundary with intent adjustment and use a strategy such as the potential hybrid control for position investigated in our previous work. This will provide the operator with partial visual feedback and partial haptic feedback. In this case, the operator will still close the loop and have full control over the system, but will not be distracted by the guidance force.

\section{Conclusion}

The presented haptic assistance adjusted based on the system's confidence in the gaze-derived operator's intent for teleoperation increases the control performance in teleoperation. It is natural and easy to use because it takes advantage of a natural characteristic of the operator's behavior. It prevents collateral damage through the use of a safety boundary which also helps the operator approach the correct depth. The results in section \ref{sec:experimental_results} show that users are faster and more accurate when using this system.

Future work will involve increasing the accuracy of the inferred intent and the confidence in this intent. This necessarily requires more information than the operator's eye movements alone. The reason for this is that the gaze is really an observational mechanism and was not intended to be a control input. In order for control to be truly natural, the operator's gaze has to be used in such a way as to not interrupt their regular behavior. However, the operator will not just look at the target during completion of the teleoperation task. He will also look at the robot end-effector, at the surrounding obstacles, or other distractions depending on the environment. Separating these eye movements, which have little to do with the final goal of the teleoperation task, from the ``valid'' fixations on the target position is very difficult without additional information. This is especially true in a more general setting where the tasks may not be related to reaching. Therefore, eye-movement data alone is likely not sufficient for a highly accurate determination of the operator's intent. 

Introducing context into the intent inference process could go a long way to solving the aforementioned issues. For example, since the location of the robot hand is known, fixations on the robot end-effector can automatically be filtered from the intent inference process. Additional steps may be to consider the actual structure of the environment being gazed at. If there is no object at the fixation location, or the object is not graspable, then the gazed-at location must not be the intended goal of the action. Furthermore, taking into account gaze history could provide additional insight into the true intent of the operator. These considerations will improve the intent inference, confidence level, and overall control significantly.

\bibliographystyle{IEEEtran}
\bibliography{GazeRobot.bib}

\end{document}